%% file: main.tex
\documentclass[runningheads]{llncs}

\usepackage{eccv}

\usepackage{eccvabbrv}

\usepackage{graphicx}
\usepackage{booktabs}

\usepackage[accsupp]{axessibility}  % Improves PDF readability for those with disabilities.

\usepackage{hyperref}

\usepackage{orcidlink}

\usepackage[colorinlistoftodos,prependcaption,textsize=tiny]{todonotes}

\usepackage{colortbl}

\usepackage{caption}

\usepackage{wrapfig}

\usepackage{floatrow}
\usepackage{marvosym}
\newfloatcommand{capbtabbox}{table}[][\FBwidth]

\begin{document}

\title{CityGaussian: Real-time High-quality Large-Scale Scene Rendering with Gaussians}

\titlerunning{CityGaussian}

\author{Yang Liu\inst{1,2}\orcidlink{0009-0005-7193-0795} \and
He Guan \inst{1,2}\orcidlink{0000-0002-8827-5935} \and
Chuanchen Luo \inst{4}\orcidlink{0000-0002-4360-7035} \and
Lue Fan \inst{1,2}\orcidlink{0000-0002-2349-0538} \and
Naiyan Wang \inst{5}\orcidlink{0000-0002-0526-3331} \and \\
Junran Peng \inst{6}\orcidlink{0000-0001-5276-0114} \textsuperscript{\Letter} \and
Zhaoxiang Zhang \inst{1,2,3} \textsuperscript{\Letter}
}

\authorrunning{Y. Liu et al.}


\institute{\textsuperscript{\rm 1} NLPR, MAIS, Institute of Automation, Chinese Academy of Sciences \\
\textsuperscript{\rm 2} University of Chinese Academy of Sciences\\
\textsuperscript{\rm 3}Centre for Artificial Intelligence and Robotic \, \textsuperscript{\rm 4}Shandong University\\
\textsuperscript{\rm 5}TuSimple \, \textsuperscript{\rm 6}University of Science and Technology Beijing\\
\email{\{liuyang2022, lue.fan, zhaoxiang.zhang\}@ia.ac.cn, chuanchen.luo@sdu.edu.cn, \{sylcito,winsty\}@gmail.com, jrpeng4ever@126.com}
}

\maketitle

\input{sections/0_abstract}
\input{sections/1_introduction}

\input{sections/2_related_works}
\input{sections/3_method}
\input{sections/4_experiments}
\input{sections/5_conclusions}

\section*{Acknowledgements}
This work was supported in part by the National Key R\&D Program of China (No. 2022ZD0116500), the National Natural Science Foundation of China (No. U21B2042, No. 62320106010), and in part by the 2035 Innovation Program of CAS, and the InnoHK program.

\bibliographystyle{splncs04}
\bibliography{egbib}

\newpage
\section*{Supplementary Material}
\input{sections/6_appendix_arXiv}

\end{document}

%% file: sections/0_abstract.tex
\begin{abstract}
  The advancement of real-time 3D scene reconstruction and novel view synthesis has been significantly propelled by 3D Gaussian Splatting (3DGS). However, effectively training large-scale 3DGS and rendering it in real-time across various scales remains challenging. This paper introduces CityGaussian (CityGS), which employs a novel divide-and-conquer training approach and Level-of-Detail (LoD) strategy for efficient large-scale 3DGS training and rendering. Specifically, the global scene prior and adaptive training data selection enables efficient training and seamless fusion. Based on fused Gaussian primitives, we generate different detail levels through compression, and realize fast rendering across various scales through the proposed block-wise detail levels selection and aggregation strategy. Extensive experimental results on large-scale scenes demonstrate that our approach attains state-of-the-art rendering quality, enabling consistent real-time rendering of large-scale scenes across vastly different scales. Our project page is available at \href{https://dekuliutesla.github.io/citygs/}{https://dekuliutesla.github.io/citygs/}.
  \keywords{Large-Scale Scene Reconstruction \and Novel View Synthesis \and 3D Gaussian Splatting}
\end{abstract}

%% file: sections/1_introduction.tex
\section{Introduction}
\label{sec:intro}

\begin{figure}
\centering
\includegraphics[width=0.99\textwidth]{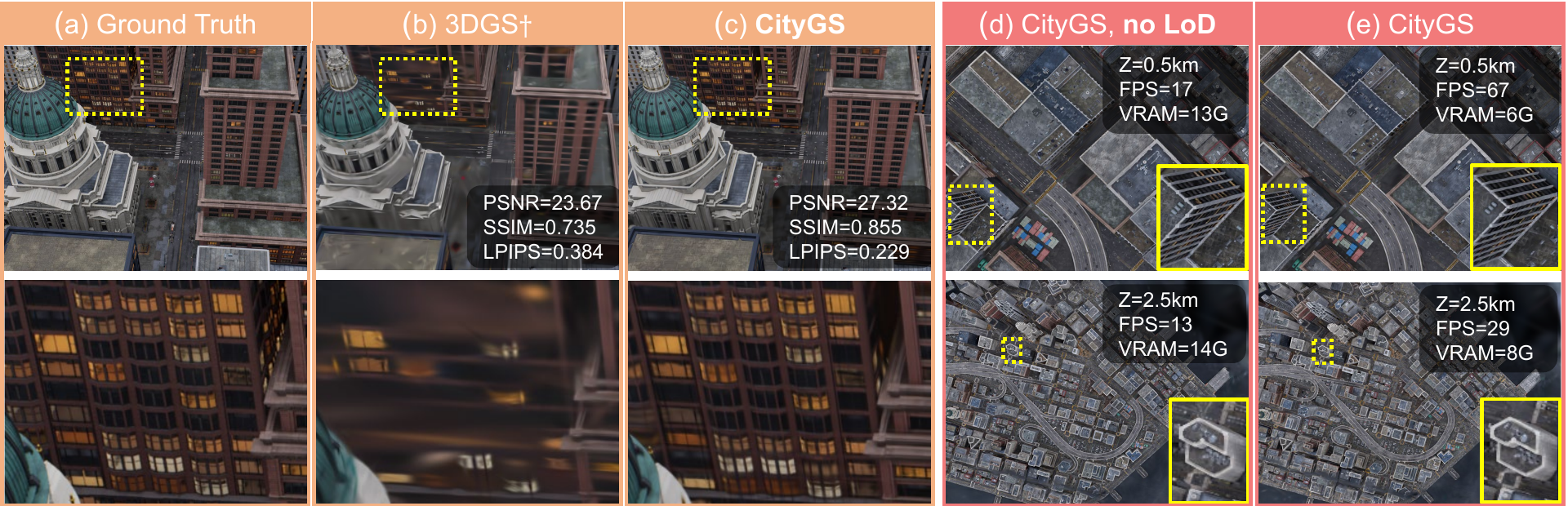}
\caption{(a, b, c) Our proposed CityGS achieves the SOTA rendering fidelity on Small City scene (5620 training images, 740 test images) of \textit{MatrixCity} dataset. The setting of baseline 3DGS$^\dagger$ is discussed in \cref{subsec:4.1-Setup}. (d, e) Here Z denotes camera height. Without LoD, CityGS would render over 20 million points, leading to considerable VRAM and time costs. The LoD saves VRAM and enables real-time performance under various scales. Note that the FPS is tested with CUDA synchronization for objective evaluation.}
\label{fig:teaser}
\end{figure}

3D large-scale scene reconstruction, as a pivotal component in AR/VR \cite{gu2023ue4,chen2024survey}, aerial surveying \cite{turki2022mega}, smart city \cite{chaturvedi2016integrating, dong2018learning}, and autonomous driving \cite{tancik2022block}, has drawn extensive attention from academia and industry in recent decades.
Such a task pursues high-fidelity reconstruction and real-time rendering at different scales for large areas that typically span over $1.5$ $km^2$ \cite{turki2022mega}. In the past few years, this field has been dominated by neural radiance fields (NeRF) \cite{mildenhall2021nerf} based methods. Representative works include Block-NeRF \cite{tancik2022block}, BungeeNeRF\cite{xiangli2022bungeenerf}, and ScaNeRF \cite{wu2023scanerf}. But they still lack fidelity in details or exhibit sluggish performance.

Recently, 3D Gaussian Splatting (3DGS) \cite{kerbl20233d} emerged as a promising alternative solution.
In contrast to NeRF, it employs explicit 3D Gaussians as primitives to represent the scene. 
% The well-optimized rasterizer endows 3DGS with rendering speed far beyond the NeRF series and comparable high quality. 
Thanks to highly efficient rasterization algorithm, 3DGS achieves high-quality visual effects at real-time rendering speed.
Most existing academic exploration around 3DGS mainly focuses on objects or small scenes. However, devils emerge when 3DGS is applied to large-scale scene reconstruction. \emph{On the one hand}, directly deploying 3DGS to large-scale scenes results in prohibitive overhead in GPU memory during training. For instance, the 24G RTX3090 raises out-of-memory errors when the Gaussian number grows above 11 million. But to reconstruct over $1.5$ $km^2$ city area with high visual quality from aerial view, over 20 million Gaussians might be required. 3DGS of such capacity can't be directly trained even on 40G A100. \emph{On the other hand}, the rendering speed bottleneck lies in depth sorting. As the number of Gaussians rises to the order of millions, the rasterization becomes extremely slow. For instance, a small \textit{Train} scene of Tanks\&Temples \cite{knapitsch2017tanks} dataset of 1.1 million Gaussians is rendered with an average visible Gaussians number of around 0.65 million and a speed of 103 FPS. But the $2.7$ $km^2$ \textit{MatrixCity} scene of 23 million Gaussians can only be rendered at the speed of 21 FPS even though the average visible Gaussians number is also around 0.65 million. And how to free unnecessary Gaussians from rasterization is the key to real-time large-scale scene rendering.

To address the problems mentioned above, we propose the \textbf{CityGaussian (CityGS)}. Inspired by MegaNeRF \cite{turki2022mega}, we adopt a divide-and-conquer strategy. The whole scene is first partitioned into spatially adjacent blocks and parallelly trained. Each block is represented by much fewer Gaussians and trained with less data, the memory cost of a single GPU is well attenuated. For Gaussian partitioning, we contract unbounded regions to normalized bounded cubic and apply uniform grid division, to better balance workloads across different blocks. For training data partitioning, a pose is kept only if it is inside the considered block, or if the considered block has a considerable contribution to the rendering result. This novel strategy efficiently avoids distraction from irrelative data, while enabling higher fidelity with less Gaussian consumption, as depicted in \cref{tab:Ablation-Training}. To align trained Gaussians from adjacent blocks, we guide the training of each block with a coarse global Gaussian prior. This strategy can efficiently avoid interactive interference among blocks and enable seamless fusion. 

To alleviate the computation burden when rendering large-scale Gaussians, we propose a block-wise Level-of-Detail (LoD) strategy. The key idea is to only feed necessary Gaussians to the rasterizer while eliminating extra computation costs. Specifically, we take the previously divided blocks as units to quickly decide which Gaussians are likely to be contained in the frustum. Furthermore, due to the perspective effect, distant regions occupy a small area of screen space and contain fewer details. Thus the Gaussian blocks that are far from the camera can be replaced with the compressed version, using fewer points and features. In this way, the processing burden of the rasterizer is significantly reduced, while the introduced extra computation remains acceptable. As illustrated in (d,e) of \cref{fig:teaser}, our CityGS can maintain real-time large-scale scene rendering even under drastically large field of view.

In a nutshell, this work has three-fold contributions:
\begin{itemize}
    \item[$\bullet$] We propose an effective divide-and-conquer strategy to reconstruct large-scale 3D Gaussian Splatting in parallel manner.
    \item[$\bullet$] With the proposed LoD strategy, we realize real-time large-scale scene rendering under drastically different scales with minimal quality loss.
    \item[$\bullet$] Our method, termed as CityGS, performs favorably against current state-of-the-art methods in public benchmarks.
\end{itemize}

%% file: sections/2_related_works.tex
\section{Related Works}

\subsection{Neural Rendering}
\subsubsection{Neural Radiance Field} is an instrumental technique for 3D scene reconstruction and novel view synthesis. 
% Neural radiance fields (NeRFs) \cite{mildenhall2021nerf}, which applies Multilayer Perceptrons (MLPs) and volumetric rendering for 3D scene representation, is instrumental in driving research and advancing the accuracy of 3D scene reconstruction and novel view synthesis. 
As an implicit neural scene representation, it employs Multilayer Perceptrons (MLPs) as the mapping function between query positions and corresponding radiances.
Volumetric rendering is then applied to render such a representation to 2D images.
The success of NeRF has spawned a wide range of follow-up works \cite{barron2021mip, barron2022mip, niemeyer2022regnerf, xu2022sinnerf, martin2021nerf, mildenhall2022nerf, pumarola2021d, reiser2023merf, tancik2023nerfstudio} that improve upon various aspects of the original method.
Among them, Mip-NeRF360 \cite{barron2022mip} serves as a recent milestone with outstanding rendering quality. However, NeRFs suffer from intensive sampling along emitted rays, resulting in relatively high training and inference latency. A series of methods \cite{Yu_Li_Tancik_Li_Ng_Kanazawa_2021, chen2022tensorf, takikawa2022variable, muller2022instant, fridovich2022plenoxels} have been proposed to alleviate this problem. And the most representatives include InstantNGP \cite{muller2022instant} and Plenoxels \cite{fridovich2022plenoxels}. Combined with a multiresolution hash grid and a small neural network, InstantNGP achieves speedup of several orders of magnitude while maintaining high image quality. On the other hand, Plenoxels represents the continuous density field with a sparse voxel grid, to get considerable speedup and outstanding performance together.

\subsubsection{Point-based Rendering}
Another parallel line of works renders the scene with point-based representation. Such explicit geometry primitives enable fast rendering speed and high editability. Pioneering works include \cite{zwicker2001ewa, lassner2021pulsar, Xu_Xu_Philip_Bi_Shu_Sunkavalli_Neumann_2022, wiles2020synsin, Yifan_Serena_Wu_Öztireli_Sorkine-Hornung_2019}, but discontinuity in rendered images remains a problem. The recently proposed 3D Gaussian Splatting (3DGS) \cite{kerbl20233d} solved this problem by using 3D Gaussians as primitives. Combined with a highly optimized rasterizer, the 3DGS can achieve superior rendering speed over the NeRF-based paradigm with no loss of visual fidelity. Nevertheless, the explicit millions of Gaussians with high dimensional features (RGB, spherical harmonics, etc) lead to significant memory and storage footprint. To mitigate this burden, methodologies such as \cite{fan2023lightgaussian, lee2023compact, morgenstern2023compact} are proposed. Apart from vector quantization \cite{navaneet2023compact3d}, \cite{fan2023lightgaussian} and \cite{morgenstern2023compact} further combine distillation and 2D grid decomposition respectively to compress storage while improving speed. Joo et al. \cite{lee2023compact} realize similar performance by exploiting the geometrical similarity and local feature similarity of Gaussians. Despite the success in compression and speedup, these researches concentrate on small scenes or single objects. Under large-scale scenes, the excessive memory cost and computation burden lead to difficulty in high-quality reconstruction and real-time rendering. And our CityGS serves as an efficient solution for these problems.

\subsection{Large Scale Scene Reconstruction}
3D reconstruction from large image collections has been an aspiration for many researchers and engineers for decades. Photo Tourism \cite{snavely2006photo} and Building Rome in a Day \cite{agarwal2011building} are two representatives of early exploration in robust and parallel large-scale scene reconstruction. With the advance of NeRF \cite{mildenhall2021nerf}, the paradigm has shifted. In Block-NeRF \cite{tancik2022block} and Mega-NeRF \cite{turki2022mega}, the divide-and-conquer strategy is adopted and each divided block is represented by a small MLP. Switch-NeRF \cite{zhenxing2022switch} further improves the performance by introducing learnable scene decomposition strategy. Urban Radiance Fields \cite{rematas2022urban} 
 and SUDS \cite{turki2023suds} further explores the large scene reconstruction with modalities beyond RGB images, such as LiDAR and 2D optical flow. To better balance the model storage and performance, Grid-NeRF \cite{xu2023grid} introduces guidance from the multiresolution feature planes, while GP-NeRF \cite{zhang2023efficient} and City-on-Web \cite{song2023city} utilizes hash grid and multiresolution tri-plane representations. To realize real-time rendering, UE4-NeRF \cite{gu2023ue4} transforms the divided sub-NeRF to polygonal meshes and combines the matured rasterization pipeline in Unreal Engine 4. ScaNeRF \cite{wu2023scanerf} further counteracts the crux of inaccurate camera poses in realistic scenes. VastGaussian \cite{lin2024vastgaussian} explores the application of 3DGS under large-scale scenes and deals with appearance variation. However, the low rendering speed of such a large scene remains a problem. In contrast, our approach combines 3D Gaussian primitives with well-designed training and LoD methodology, elevating real-time render quality by a large margin.

\subsection{Level of Detail}
In computer graphics, Level of Detail (LoD) techniques regulate the amount of detail used to represent the virtual world so as to bridge complexity and performance \cite{luebke2003level}. Typically, LoD decreases the workload of objects that are becoming less important (e.g. moving away from the viewer). In recent years, the incorporation of LoD and the neural radiance field has received extensive concern. BungeeNeRF \cite{xiangli2022bungeenerf} equips NeRF with a progressive growing strategy, where each residual block is responsible for a finer detail level. On the other hand, NGLoD \cite{takikawa2021neural} represents different detail levels of neural signed distance functions (SDFs) with multi-resolution sparse voxel octree, while VQ-AD \cite{takikawa2022variable} uses hierarchical feature grid to compactly represent LoD of 3D signals. Motivated by Mip-NeRF \cite{mildenhall2021nerf}, both Tri-MipRF and LoD-NeuS \cite{zhuang2023anti} apply cone casting with multiresolution tri-plane representation for anti-aliased LoD. City-on-Web \cite{song2023city} generates coarser detail levels by training feature grid of lower resolution and corresponding deferred MLP decoder. By realizing LoD on explicit 3D Gaussian representation, we efficiently improve its real-time performance under large-scale scenes.

%% file: sections/3_method.tex
\section{Method}
\textbf{Overview} \quad The training and rendering pipelines are respectively shown in \cref{fig:train} and \cref{fig:render}. We first generate a 3DGS that offers global scene depiction with normal 3DGS training strategy described in \cref{sec:3.1-Preliminary}.  Building upon this global prior, we employ the training strategy presented in \cref{sec:3.2-Training} to adaptively divide the Gaussian primitives and data for further parallel training. Based on the fused large-scale Gaussians, our Level of Detail (LoD) algorithm discussed in \cref{sec:3.3-LoD} dynamically selects required Gaussians for fast rendering.

\subsection{Preliminary}
\label{sec:3.1-Preliminary}
We begin with a brief introduction of 3DGS \cite{kerbl20233d}. The 3DGS represents the scene with discrete 3D Gaussians $\mathbf{G}_{\mathbf{K}}=\left\{ G_k|k=1,...,K \right\}$, each of them is equipped with learnable properties including 3D position $\boldsymbol{p}_{\boldsymbol{k}}\in \mathbb{R} ^{3\times 1}  $, opacity $\alpha _k\in \left[ 0,1 \right] $, geometry (i.e. scaling and rotation used to construct Gaussian covariance), spherical harmonics (SH) features $\boldsymbol{f}_{\boldsymbol{k}}\in \mathbb{R} ^{3\times 16}$ for view-dependent color $\boldsymbol{c}_{\boldsymbol{k}}\in \mathbb{R} ^{3\times 1}$. In rendering, given the intrinsics $\kappa$ and pose $\tau_i$ of $i$-th image, the Gaussians are splatted to screen space, sorted in depth order, and rendered via alpha blending:
\begin{equation}
    \label{eq:3-alpha-blending}
     c_i\left( \boldsymbol{x} \right) =\sum_{k=1}^K{\alpha _k\boldsymbol{c}_{\boldsymbol{k}}G_{k}^{2D}\left( \boldsymbol{x} \right)}\prod_{t=1}^{k-1}{\left( 1-\alpha _tG_{t}^{2D}\left( \boldsymbol{x} \right) \right)}, G_{k}^{2D}=\mathrm{\textbf{proj}}\left(G_k, \kappa, \tau_i\right),
\end{equation}
where $\mathrm{\textbf{proj}}$ is projection operation, $c_i\left( \boldsymbol{x} \right)$ is the color at pixel position $\boldsymbol{x}$, $G_{k}^{2D}$ is projected Gaussian distribution. We refer the readers to the original paper \cite{kerbl20233d} for details. The final rendered image is denoted as $I_{\mathbf{G}_{\mathbf{K}}}\left( \tau _i \right)$. During training, a typical initialization choice is the point cloud generated with the Structure-from-Motion (SfM), such as COLMAP \cite{schoenberger2016sfm}. Then based on gradients derived from differentiable rendering, the Gaussians would be cloned, densified, pruned, and consistently refined. However, the depiction of large-scale scenes can consume over 20 million primitives, which easily causes out-of-memory errors in training process, and rendering time is slowed down as well.

\subsection{Training CityGS}
\label{sec:3.2-Training}

\begin{figure}
\centering
\includegraphics[height=2.35cm]{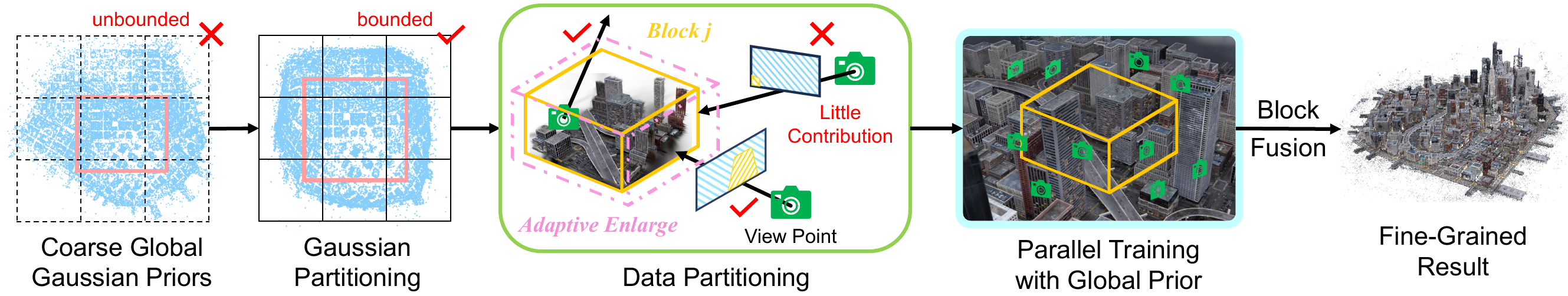}
\caption{The training process of CityGS. The \textcolor[RGB]{255,159,159}{pink} square bounds the foreground area, facilitating subsequent unbounded space contraction and Gaussian partitioning. Then for a specific block, a pose is assigned to training set if it is inside the block or if the block has a considerable contribution to the rendering result. These blocks are then parallelly trained and merged togethor to depict the whole scene.}
\label{fig:train}
\end{figure}

In this section, we first illustrate the necessity of a coarse global Gaussian prior and how to generate it. Based on this prior, we describe the Gaussian and data primitives division strategy. At the end of this section, we present the training and post-processing details. The pipeline is shown in \cref{fig:train}.

\subsubsection{Global Gaussian Prior Generation}
\label{subsec:3.2.1-Global Gaussian}
This part serves as the basis for further Gaussian and data division. An intuitive large-scale scene training strategy involves applying a divide-and-conquer strategy to the COLMAP points. However, due to a lack of depth constraints and global awareness, many geometrically inaccurate floaters will be generated to overfit regions outside the block, making a reliable fusion of different blocks difficult. Additionally, the rendered images from COLMAP points tend to be blurred and inaccurate, making it difficult to assess whether a specific view is important to training the block. To this end, we propose a simple yet effective way to solve the problems. Specifically, we first train the COLMAP points with all observations for 30,000 iterations, yielding a coarse description of the overall geometry distribution. The resulting set of Gaussian primitives are denoted as $\mathbf{G}_{\mathbf{K}}=\left\{ G_k|k=1,...,K \right\}  $, where $K$ is the total amount. In further block-wise finetuning, such a strong global geometry prior leads point to geometrically appropriate positions, eliminating severe interference in fusion. As proved in \cref{tab:Ablation-Training}, this strategy can efficiently improve rendering fidelity. Moreover, this coarse Gaussian provides more accurate geometry distribution and cleaner rendered images, facilitating subsequent primitives and data division.

\subsubsection{Primitives and Data Division}
\label{subsec:3.2.2-Division}

% Under normally adopted training view distribution, the central part of the considered region receives more supervision and is thus represented by dense Gaussians. In contrast, the boundary regions are sparsely supervised, and thus are occupied by sparsely and randomly distributed Gaussians. A direct Gaussian partitioning according to a uniform grid will lead to considerably uneven Gaussian assignment. For instance, in the leftmost section of \cref{fig:train}, the left bottom grid is assigned with a few Gaussians, while the central grid is densely occupied. As a result, under the same training schedule, the boundary blocks prone to overfit the sparse supervision, while the central blocks may underfit the dense supervision. To address this issue, we first contract the Gaussians to a bounded cubic region.

Considering most real-world scenes are unbounded \cite{tancik2023nerfstudio}, the optimized Gaussians could extend indefinitely. Directly applying uniform grid division on the original 3D space would lead to many almost empty blocks, thus making workloads severely unbalanced. To alleviate this problem, we first contract the global Gaussian prior to a bounded cubic region. 

For contraction, the inner foreground region, i.e. the \textcolor[RGB]{255,159,159}{pink} square in \cref{fig:train}, contains a linear space mapping, while the outer background region contains a non-linear mapping. Specifically, we depict the foreground region with the minimum and maximum positions of its corners $\boldsymbol{p}_{\min}$ and $\boldsymbol{p}_{\max}$. Then we normalize Gaussian positions as $\hat{\boldsymbol{p}}_{\boldsymbol{k}}=2\left( \boldsymbol{p}_{\boldsymbol{k}}-\boldsymbol{p}_{\min} \right) /\left( \boldsymbol{p}_{\max}-\boldsymbol{p}_{\min} \right) - 1 $. Consequently, the positions of foreground Gaussians fall within the range $[-1, 1]$. The subsequent contraction is executed using the following function \cite{wu2023scanerf}:
\begin{equation}
\mathrm{\textbf{contract}}\left( \hat{\boldsymbol{p}}_{\boldsymbol{k}} \right) =\begin{cases}
	\hat{\boldsymbol{p}}_{\boldsymbol{k}},                              &\text{if}\,\,||\hat{\boldsymbol{p}}_{\boldsymbol{k}}||_{\infty}\leqslant 1,\\
	\left( 2-\frac{1}{||\hat{\boldsymbol{p}}_{\boldsymbol{k}}||_{\infty}} \right) \frac{\hat{\boldsymbol{p}}_{\boldsymbol{k}}}{||\hat{\boldsymbol{p}}_{\boldsymbol{k}}||_{\infty}},   &\text{if}\,\,||\hat{\boldsymbol{p}}_{\boldsymbol{k}}||_{\infty}>1.\\
\end{cases}
\end{equation}
By evenly partitioning this contracted cubic space, a more balanced Gaussian partitioning is derived. 

In the finetuning phase, we hope each block is sufficiently trained. To be specific, the assigned data should keep training progress focused on refining details within the block. 
So one pose needs to be retained only when the considered block projects a significant amount of visible content to the rendering results. Distractive cases like severe occlusion or minor content contribution should be occluded. Since SSIM loss can efficiently capture the structural difference and is to some extent insensitive to brightness changes \cite{wang2003multiscale}, we take it as the basis of our data partition strategy. 

Specifically, for the $j$-th block, the containing coarse global Gaussians are denoted as $\mathbf{G}_{\mathbf{K}_{\mathbf{j}}}=\left\{ G_k|\mathbf{b}_{\mathbf{j},\mathbf{min}}\leqslant \mathrm{\textbf{contract}}\left( \hat{\boldsymbol{p}}_{\boldsymbol{k}} \right) <\mathbf{b}_{\mathbf{j},\mathbf{max}}, k=1,...,K_j \right\} $, where $\mathbf{b}_{\mathbf{j,min}}$ and $\mathbf{b}_{\mathbf{j,max}}$ defines the x,y,z bound of block $j$, and $K_j$ is the number of contained Gaussians. Then whether the $i$-th pose $\tau _i$ is assigned to $j$-th block is determined by:
\begin{equation}
    \boldsymbol{B}_1\left( \tau _i, \mathbf{G}_{\mathbf{K}_{\mathbf{j}}} \right) =\begin{cases}
	1, \quad L_{\mathrm{SSIM}}\left( I_{\mathbf{G}_{\mathbf{K}}}\left( \tau _i \right) , I_{\mathbf{G}_{\mathbf{K}}\setminus\mathbf{G}_{\mathbf{K}_{\mathbf{j}}}}\left( \tau _i \right) \right) >\varepsilon,\\
	0, \quad \text{otherwise},\\
\end{cases}
\end{equation}
where $\mathbf{G}_{\mathbf{K}}\setminus\mathbf{G}_{\mathbf{K}_{\mathbf{j}}}$ defines difference set of $\mathbf{G}_{\mathbf{K}}$ and $\mathbf{G}_{\mathbf{K}_{\mathbf{j}}}$. The SSIM loss $L_{\mathrm{SSIM}}$ larger than threshold $\varepsilon$ means a considerable contribution of block $j$ to the rendered image and thus leads to an assignment. 

However, solely relying on the first principle can lead to artifacts when viewing outside at the edge of the block. Because these cases rarely involve the projection of considered block, they will not be sufficiently trained under the first principle. Therefore we also include poses that fall into considered blocks, i.e.
\begin{equation}
    \boldsymbol{B}_2\left( \tau _i, \mathbf{G}_{\mathbf{K}_{\mathbf{j}}} \right) =\begin{cases}
	1,\quad\mathbf{b}_{\mathbf{j},\mathbf{min}}\leqslant \mathrm{\textbf{contract}}\left( \hat{\boldsymbol{p}}_{\boldsymbol{\tau }_{\boldsymbol{i}}} \right) <\mathbf{b}_{\mathbf{j},\mathbf{max}},\\
	0, \quad \text{otherwise},\\
\end{cases}
\end{equation}
where $\hat{\boldsymbol{p}}_{\boldsymbol{\tau }_{\boldsymbol{i}}}$ is the position under world coordinate of pose $i$. And the final assignment is:
\begin{equation}
    \boldsymbol{B}\left( \tau _i, \mathbf{G}_{\mathbf{K}_{\mathbf{j}}} \right) =\boldsymbol{B}_1\left( \tau _i, \mathbf{G}_{\mathbf{K}_{\mathbf{j}}} \right) + \boldsymbol{B}_2\left( \tau _i, \mathbf{G}_{\mathbf{K}_{\mathbf{j}}} \right).
\end{equation}

Despite having the above strategies in place, empty blocks may still exist in cases of extremely uneven distributions or high block dimensions. To prevent overfitting, we enlarge the bound $\mathbf{b}_{\mathbf{j,min}}$ and $\mathbf{b}_{\mathbf{j,max}}$ until $K_j$ exceeds certain threshold. This process is only used in data assignment to ensure enough training data for each block. 

\subsubsection{Finetuning and Post-processing} 
\label{subsec:3.2.3-Training&Post-processing}
After data and primitives division, we proceed to train each block in parallel. \textbf{It is worth noting this fine-tuning stage is under original uncontracted space}. Specifically, we utilize the coarse global prior generated in \cref{subsec:3.2.1-Global Gaussian} to initialize the finetuning of each block. The training loss follows the approach outlined in the original 3DGS paper \cite{kerbl20233d}, comprising a weighted sum of $L1$ loss and SSIM loss. Then for each block, we filter out the finetuned Gaussians contained within its spatial bound. Thanks to the global geometric prior, interference among blocks is significantly mitigated. Thus a high-quality overall model can be derived through direct concatenation. Additional qualitative validation can be found in the Appendix. Further refinement is left in the LoD part.

\begin{figure}
\centering
\includegraphics[width=0.99\textwidth]{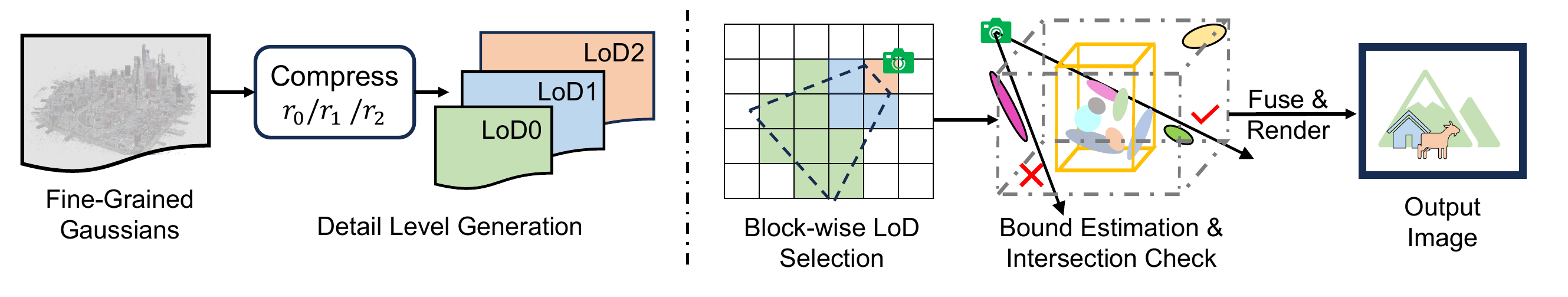}
\caption{Rendering of CityGS. Based on trained dense Gaussians, we generate detail levels with different compression rates $r_0$, $r_1$ and $r_2$. When rendering, all the Gaussians in the same block will share the same detail level, which is determined by the block's distance to the camera. Since the contraction-based block partition leads to some irregular block shapes, we estimate their bounding boxes after removing floaters. The frustum intersection with the estimated block shape determines whether the block will be fed to rasterizer.}
\label{fig:render}
\end{figure}

\subsection{Level-of-Detail on CityGS}
\label{sec:3.3-LoD}

As discussed in \cref{sec:intro}, to eliminate the computation burden brought by unnecessary Gaussians to rasterizer, our CityGS involves multiple detail levels generation and block-wise visible Gaussian selection. We introduce these two parts respectively in the following two subsections.

\subsubsection{Detail Level Generation}
\label{subsubsec:3.3.1-Generation}
As objects move away from the camera, they occupy less area on the screen, while contributing less high-frequency information. Thus distant, low detail-level regions can be well represented by models of low capacity, i.e. fewer points, less feature dimension, and lower data precision. In practice, we generate different detail levels with the advancing compression strategy LightGaussian \cite{fan2023lightgaussian}, which operates directly on trained Gaussians, achieving a substantial compression rate with minimal performance degradation. Consequently, the memory and computation demand for required Gaussians is significantly alleviated, while the rendering fidelity is still well maintained.

% The models of the finest several levels are usually stored on the CPU, and only necessary parts are passed to the GPU for rendering.

\subsubsection{Detail Level Selection and Fusion}
\label{subsubsec:3.3.2-Selection}
A baseline of detail level selection is to fill frustum regions between different distance intervals with Gaussians from the corresponding detail level. However, this method necessitates per-point distance calculation and assignment, resulting in significant computational overhead, as confirmed in \cref{subsec:4.4-Ablation}. Therefore, we adopt a block-wise strategy, considering spatially adjacent blocks as the unit, as depicted in the right part of \cref{fig:render}. Each block is considered as a cubic with eight corners for calculation of frustum intersection. All the Gaussians contained in certain blocks will share the same detail level, which is determined by the minimum distance from eight corners to the camera center. However, in practice, we found the minimum and maximum coordinates of Gaussians are usually determined by floaters. The resulting volume would be unreasonably enlarged, leading to many fake intersections. To avoid the influence of these floaters, we take the Median Absolute Deviation (MAD) \cite{dodge2008concise} algorithm. The bounds of the $j$-th block, denoted as $\boldsymbol{p}_{\mathbf{min}}^{\boldsymbol{j}}$ and $\boldsymbol{p}_{\mathbf{max}}^{\boldsymbol{j}}$, are determined by:
\begin{equation}
    \begin{aligned}
    MAD_j&=\mathrm{median}\left( \left| \boldsymbol{p}_{\boldsymbol{k}}^{\boldsymbol{j}}-\mathrm{median}\left( \boldsymbol{p}_{\boldsymbol{k}}^{\boldsymbol{j}} \right) \right| \right) ,
    \\
    \boldsymbol{p}_{\mathbf{min}}^{\boldsymbol{j}}&=\max \left( \min \left( \boldsymbol{p}_{\boldsymbol{k}}^{\boldsymbol{j}} \right) , \mathrm{median}\left( \boldsymbol{p}_{\boldsymbol{k}}^{\boldsymbol{j}} \right) -n_{MAD}\times MAD_j \right) ,
    \\
    \boldsymbol{p}_{\mathbf{max}}^{\boldsymbol{j}}&=\min \left( \max \left( \boldsymbol{p}_{\boldsymbol{k}}^{\boldsymbol{j}} \right) , \mathrm{median}\left( \boldsymbol{p}_{\boldsymbol{k}}^{\boldsymbol{j}} \right) +n_{MAD}\times MAD_j \right) ,
    \end{aligned}
\end{equation}
where $n_{MAD}$ is the hyper-parameter. By choosing the appropriate $n_{MAD}$, this method can capture the bound of the block more accurately. 

After that, all the corners in front of the camera will be projected into screen space. The minimum and maximum of these projected points composite a bounding box. By calculating its Intersection-over-Union (IoU) with screen area, we can check if the block has an intersection with the frustum. Along with the block where the camera is in, all visible blocks of corresponding detail level will be used for rendering. 

% Then we sort these blocks with distance to the camera, which is determined by the minimum distance from eight corners to the camera center. Comparing the distance with thresholds, each block will be assigned with corresponding detail level. 

In the fusion step, different detail levels are still comprehended via direct concatenation, which generates negligible discontinuity.

% \subsubsection{Data Transfer and Fusion}
% \label{subsec:3.3.3-Transfer}
% To speed up data transfer, we pass the position of Gaussians in float32, while the rest properties in float16. The fusion of different detail levels is still realized by direct concatenation, which generates negligible discontinuity.

%% file: sections/4_experiments.tex
\section{Experimets}
\label{sec:4-Exp}

\subsection{Experiments setup}
\label{subsec:4.1-Setup}

\subsubsection{Dataset and Metrics}
\label{subsubsec:4.1.1-Dataset&Metrics}

Our algorithm is benchmarked on five scenes with various scales and environments. Specifically, we adopt the 2.7$km^2$ Small City scene of synthetic city-scale dataset \textit{MatrixCity} \cite{li2023matrixcity}, as done in \cite{song2023city}. However, instead of solely training and evaluating on partial city area, we construct the whole city and compare performances. And we rescale the image width to 1600 pixels. We also carried out experiments on public real-world scene datasets, including \textit{Residence}, \textit{Sci-Art}, \textit{Rubble}, and \textit{Building} \cite{turki2022mega}. Following the approach in \cite{turki2022mega, zhenxing2022switch, zhang2023efficient}, the image resolution for these datasets is reduced by a factor of $4\times$. To further validate the generalization ability of our CityGS, we also test it on the street view scene Block\_A of \textit{MatrixCity} \cite{li2023matrixcity}. This challenging scene contains 4076 training images and 495 test images. To comprehensively measure the reconstruction quality of different methods, we take standard \textbf{SSIM}, \textbf{PSNR}, and \textbf{LPIPS} as our metrics \cite{zhang2018unreasonable}. We also compare \textbf{FPS} to evaluate the rendering speed.  It is noteworthy that we synchronize all CUDA streams before measuring time, ensuring an objective evaluation of render times for each frame.

\begin{figure}
\centering
\includegraphics[width=0.99\textwidth]{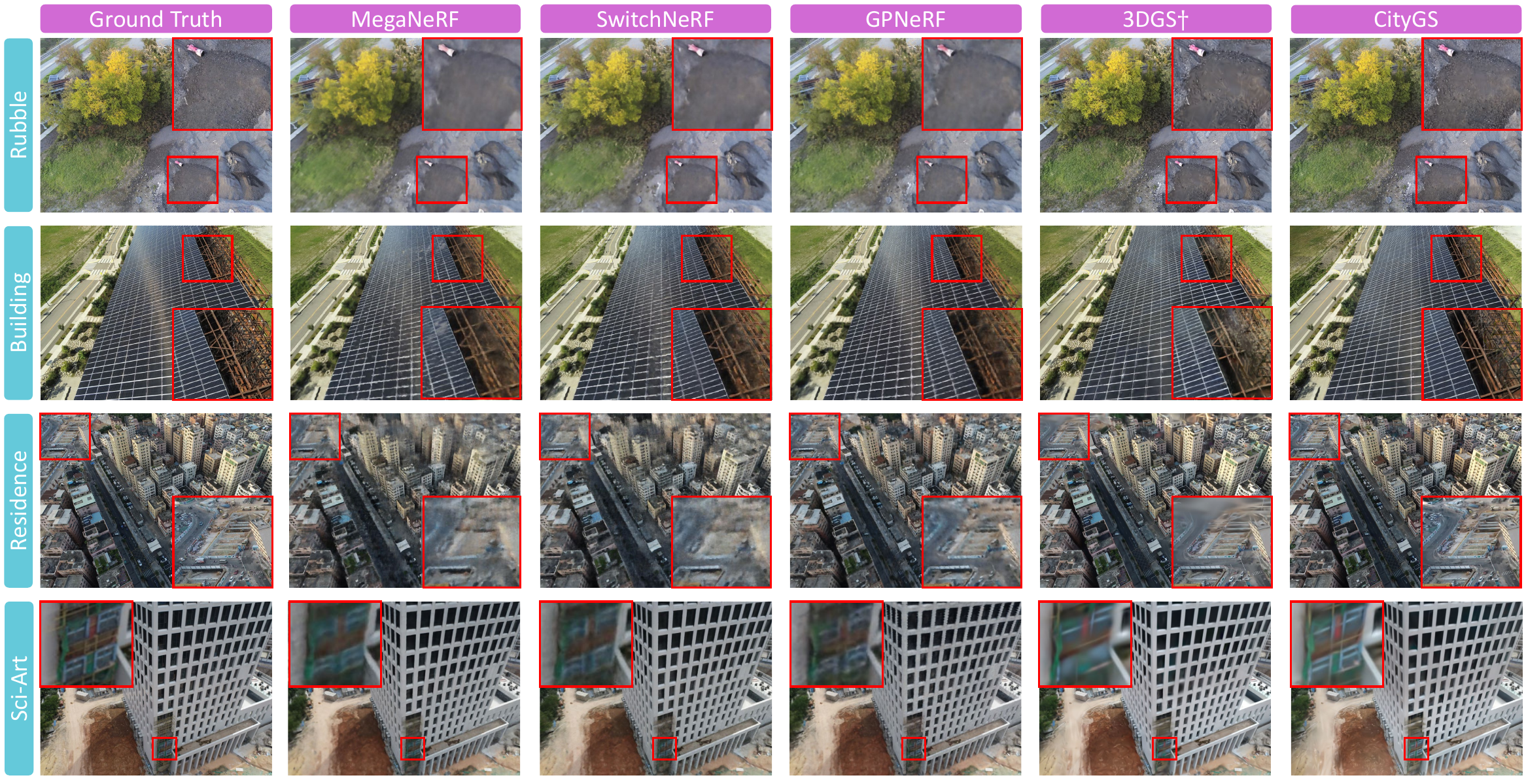}
\caption{Qualitative comparison with SOTA methods on real-scene datasets.}
\label{fig:Vis}
\end{figure}

\subsubsection{Implementations and Baselines}
\label{subsubsec:4.1.2-Implementations&Baselines}

Our method first trains coarse global Gaussian priors for further block-wise refinement. The training of priors for \textit{Sci-Art}, \textit{Residence} and \textit{Rubble} adheres to the parameter settings outlined in 3DGS \cite{kerbl20233d}. But for \textit{Building} and \textit{MatrixCity}, we halve the learning rate of position and scaling to prevent underfitting caused by aggressive optimization. In the fine-tuning phase, we train each block for another 30,000 iterations with coarse global Gaussian prior as initialization. Furthermore, the learning rate of position is reduced by 60\%, while that of scaling is empirically reduced by 20\%, compared to setting in 3DGS \cite{kerbl20233d}. The foreground area of contraction function is chosen as the central 1/3 area, and detailed block dimensions can be found in the Appendix. Our method is benchmarked against Mega-NeRF \cite{turki2022mega}, Switch-NeRF \cite{zhenxing2022switch}, GP-NeRF \cite{zhang2023efficient}, and 3DGS \cite{kerbl20233d}. Since the datasets contain thousands of images, which are much larger than that used in the original 3DGS paper \cite{kerbl20233d}, we adjust the total iteration to 60,000, while densifying Gaussians from iteration 1,000 to 30,000 with an iteration interval of 200. This strong baseline is named as 3DGS$^\dagger$. In LoD, we evaluate on the \textit{MatrixCity} dataset. We use 3 detail levels, where LoD 2 is the finest and LoD 0 is the coarsest. $n_{MAD}$ mentioned in \cref{subsubsec:3.3.2-Selection} is set to 4. The blocks within 0m to 200m are represented by LoD 2, blocks within 200m to 400m are represented by LoD 1, while others are represented by LoD 0.

\subsection{Comparison with SOTA}
\label{subsec:4.2-SOTA}
\begin{figure}
\centering
\includegraphics[width=0.99\textwidth]{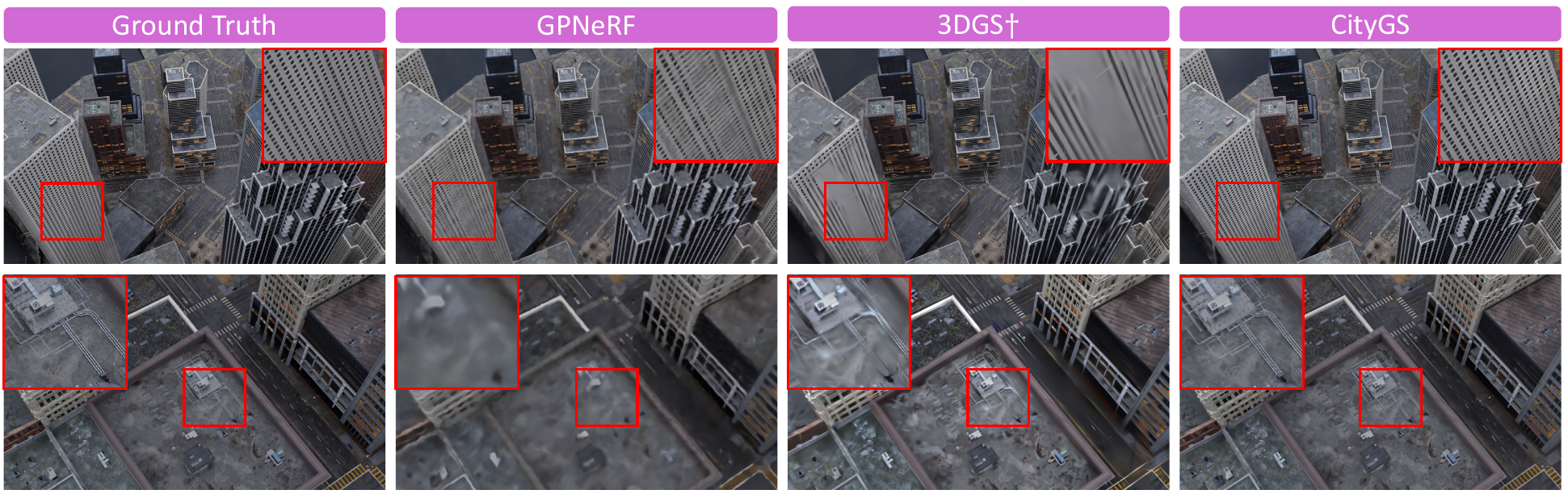}
\caption{Qualitative comparison with SOTA methods on \textit{MatrixCity} dataset.}
\label{fig:Vis-Sim}
\end{figure}

For an apple-to-apple rendering quality comparison with the SOTA large-scale scene reconstruction strategy, we use the performance of the CityGS with no LoD. The quantitative result is shown in \cref{tab:SOTA}. Due to page limits, we put results of \textit{Sci-Art} in the Appendix. It can be observed that our method outperforms the NeRF-based baselines by a large margin. On the one hand, as far as we know, our method is the first of all attempts that successfully reconstructs the whole \textit{MatrixCity} with highly variable camera altitude, ranging from 150m to 500m \cite{li2023matrixcity}. The \textbf{PSNR} arrives at 27.46 and qualitative results presented in \cref{fig:Vis-Sim}  also validate the high fidelity of our renderings. Compared with the baseline 3DGS$^\dagger$, our CityGS can capture much richer details. More visualizations on \textit{MatrixCity} can be found in the Appendix. On the other three realistic scenes, our method enables much higher \textbf{SSIM} and \textbf{LPIPS}, which indicates outstanding visual quality. As shown in \cref{fig:Vis}, thin structures like girder steel and window frames can be well reconstructed. The details of complex structures like grass and rubble are well recovered as well. In addition, results in \cref{tab:Street View} validates that our training strategy can generalize well to street scenes.

\begin{table}
  \caption{Quantitative Comparison on four large-scale scene datasets. The '-' symbol indicates Mega-NeRF \cite{turki2022mega} and Switch-NeRF \cite{zhenxing2022switch} were not evaluated on \textit{MatrixCity} due to difficulties in adjusting its training configurations beyond the provided, resulting in poor performance on this dataset. 'n/a' means the number of Guassians '\#GS' does not apply to NeRF-based methods. The best results of each metric are in \textbf{bold}.
  }
  \label{tab:SOTA}
  \centering
  \tabcolsep=0.1cm
  \resizebox{0.99\textwidth}{!}{%
  \begin{tabular}{@{}l|llll|llll|llll|llll@{}}
    \toprule
     & \multicolumn{4}{c|}{MatrixCity} & \multicolumn{4}{c|}{Residence} & \multicolumn{4}{c|}{Rubble} & \multicolumn{4}{c}{Building} \\
    \midrule
    Metrics & SSIM$\uparrow$ & PSNR$\uparrow$ & LPIPS$\downarrow$ & \#GS & SSIM$\uparrow$ \ & PSNR$\uparrow$ & LPIPS$\downarrow$ & \#GS & SSIM$\uparrow$ \ & PSNR$\uparrow$ & LPIPS$\downarrow$ & \#GS & SSIM$\uparrow$ \ & PSNR$\uparrow$ & LPIPS$\downarrow$ & \#GS\\
    \midrule
    MegaNeRF \cite{turki2022mega} & - & - & - & n/a & 0.628 & 22.08 & 0.489 & n/a & 0.553 & 24.06 & 0.516 & n/a & 0.547 & 20.93 & 0.504 & n/a \\
    Switch-NeRF \cite{zhenxing2022switch} & - & - & - & n/a & 0.654 & \textbf{22.57} & 0.457 & n/a & 0.562 & 24.31 & 0.496 & n/a & 0.579 & 21.54 & 0.474 & n/a \\
    GP-NeRF \cite{xu2023grid} & 0.611 & 23.56 & 0.630 & n/a & 0.661 & 22.31 & 0.448 & n/a & 0.565 & 24.06 & 0.496 & n/a & 0.566 & 21.03 & 0.486 & n/a \\
    3DGS$^\dagger$ \cite{kerbl20233d} & 0.735 & 23.67 & 0.384 & 9.7M & 0.791 & 21.44 & 0.236 & 6.6M & 0.777 & 25.47 & 0.2774 & 6.1M & 0.720 & 20.46 & 0.305 & 6.4M \\
    Ours & \textbf{0.865} & \textbf{27.46} & \textbf{0.204} & 23.7M & \textbf{0.813} & 22.00 & \textbf{0.211} & 10.8M & \textbf{0.813}& \textbf{25.77} & \textbf{0.228} & 9.7M & \textbf{0.778} & \textbf{21.55} & \textbf{0.246} & 13.2M\\
    \bottomrule
  
  \end{tabular}
  }
\end{table}

\subsection{Level of Detail}
\label{subsec:4.3-LoD}

\begin{figure}
\centering
\includegraphics[width=0.99\textwidth]{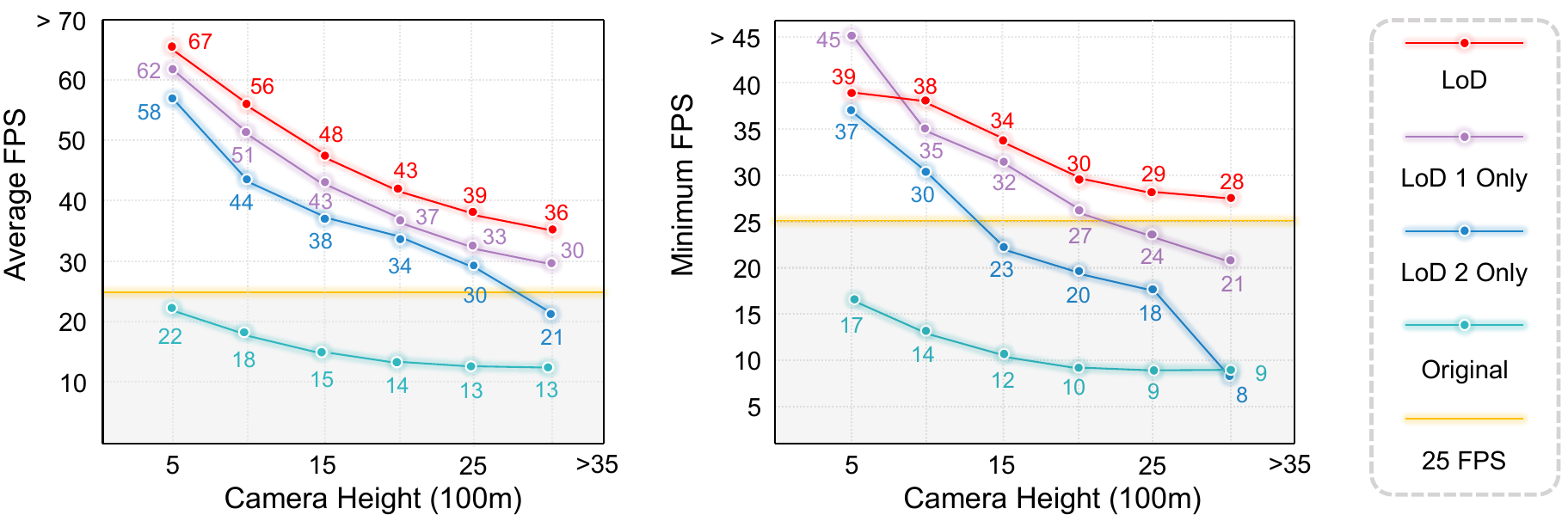}
\caption{Validation of LoD. We test rendering speed under looking-down cameras of different heights. The left part denotes \textbf{average} FPS, while the right part denotes \textbf{minumum} FPS. The shadowed area under 25 FPS line is the non-real-time zone.}
\label{fig:LoD}
\end{figure}

Considering that \textit{MatrixCity} has test split of over 700 images and various altitudes, we take it as our benchmark to evaluate effectiveness of 
 our LoD strategy. Specifically, we first generate three detail levels with compression rates 50\%, 34\%, and 25\%, namely LoD 2, LoD 1, and LoD 0. Our CityGS then applies the proposed LoD technique to combine all these detail levels. As shown in \cref{tab:LoD}, the most fine-grained LoD 2 gains the best rendering quality, while the coarsest LoD 0 has the fastest rendering speed. Compared with the three detail levels, the version with LoD technique obtains SSIM and PSNR only second to LoD 2, while the speed is very close to LoD 1. 

\begin{table}
    \caption{Validation of LoD. Without considering the result with no LoD, the best performances are in \textbf{bold}, while the second best results are in \textcolor{red}{red}. Only LoD $i$ means only using $i$-th detail level for rendering. And LoD means applying all three detail levels.}
    \label{tab:LoD}
    \centering
    \tabcolsep=0.6cm
    \resizebox{0.8\textwidth}{!}{%
    \begin{tabular}{@{}l|llll@{}}
    \toprule
    Models & SSIM$\uparrow$ & PSNR$\uparrow$ & LPIPS$\downarrow$ & FPS$\uparrow$ \\
    \midrule
    no-LoD & 0.865 & 27.46 & 0.204 & 21.6 \\
    Only LoD 2 & \textbf{0.863} & \textbf{27.54} & \textbf{0.215} & 45.6 \\
    Only LoD 1 & 0.848 & 27.20 & 0.244 & \textcolor{red}{57.2} \\
    Only LoD 0 & 0.825 & 26.57 & 0.279 & \textbf{69.4} \\
    LoD & \textcolor{red}{0.855} & \textcolor{red}{27.32} & \textcolor{red}{0.229} & 53.7 \\
    \bottomrule
  \end{tabular}}
\end{table}

% Part (a) shows the rendering results from detail level 2 only and full detail levels. From left to right, the altitudes are 0.5km, 1km, and 2.5km. In part (b)

\begin{wraptable}{r}{0.36\linewidth}
    \caption{Quantitative Comparison on \textit{MatrixCity} street scene Block\_A. The best results of each metric are in \textbf{bold}.}
    \label{tab:Street View}
    \centering
    \tabcolsep=0.1cm
    \resizebox{0.99\textwidth}{!}{
    \begin{tabular}{cccc}
    \toprule
    Method & SSIM$\uparrow$ & PSNR$\uparrow$ & LPIPS$\downarrow$ \\
    \midrule
    MipNeRF360\cite{barron2022mip} & 0.717 & 22.00 & 0.488 \\
    3DGS$^\dagger$\cite{kerbl20233d} & 0.701 & 21.14 & 0.441 \\
    % 3DGS-12W & 0.726 & 21.56 & 0.413 \\
    CityGS & \textbf{0.808} & \textbf{22.98} & \textbf{0.301} \\
    \bottomrule
    \end{tabular}}
\end{wraptable}

However, the camera altitude of \textit{MatrixCity} test split is bounded below 500m. To validate performance under extreme scale variance, we adjust the pitch to view straightly downside with appointed altitude, while keeping other pose attributes unchanged. The rendering results from different altitudes can be found in parts (d) and (e) of \cref{fig:teaser}. The visual disparity between versions with and without LoD can be observed only when zooming deeply in.

In \cref{fig:LoD}, we present how the mean and minimum rendering speed changes with different camera heights. LoD 0 is not considered for its inferior rendering quality. Comparing the mean FPS, both results from LoD and single detail level greatly outperform vanilla by a large margin. And the LoD version wins the highest speed at all heights. Comparing the minimum FPS, only the LoD version gains consistent real-time performance under worst cases of various heights, while the worst speed of single detail level drops dramatically as the camera lifts up. In other words, the proposed LoD helps smooth real-time transitions among drastically varying scales.

\subsection{Ablation}
\label{subsec:4.4-Ablation}

\begin{table}
    \caption{Ablation on training strategy. The experiment is conducted on the \textit{Rubble} dataset. The baseline denotes simply selecting cameras within the 1.5× larger area of the block for training. The best performances are in \textbf{bold}.}
    \label{tab:Ablation-Training}
    \centering
    \tabcolsep=0.4cm
    \resizebox{0.75\textwidth}{!}{%
    \begin{tabular}{@{}r|llll@{}}
    \toprule
     & SSIM$\uparrow$ & PSNR$\uparrow$ & LPIPS$\downarrow$ & \#GS \\
    \midrule
     baseline & 0.779 & 23.98 & 0.251 & 12.2 \\
     w global, baseline & 0.801 & 25.01 & \textbf{0.227} & 15.4 \\
     w/o Eq.(3), Ours & 0.749 & 23.43 & 0.252 & 19.7 \\
     w/o Eq.(4), Ours & 0.810 & 25.45 & 0.233 & \textbf{9.6} \\
     Full, Ours & \textbf{0.813} & \textbf{25.77} & 0.231 & 9.7 \\
    \bottomrule
  \end{tabular}
  }
\end{table}

In this section, we analyze the impact of training strategies as shown in \cref{tab:Ablation-Training}. As a baseline, we trained using cameras in a 1.5× larger area of each block. The results in the first and last lines of \cref{tab:Ablation-Training} indicate that this method significantly underperforms compared to our CityGS. The main issue is floaters caused by overfitting areas outside the block, but this is mitigated with guidance from the global model, improving performance as seen in the second line of \cref{tab:Ablation-Training}. However, this improved baseline uses 1.5× more Gaussians than CityGS, making it inefficient. The last three lines demonstrate that Eq.(3) is crucial for data partitioning, and we found Eq.(4) helps prevent floaters at block edges. Further hyper-parameter ablations are detailed in the Appendix.

\begin{table}
  \caption{Ablation on LoD Strategy. In-Frustum denotes the strategy used to select Gaussians within frustum. The block-wise and point-wise strategy are discussed in \cref{subsubsec:3.3.2-Selection}. Distance Interval here denotes the distance interval used by different detail levels. The area within the near interval would be represented by a higher detail level. The best results are in \textbf{bold}.}
  \label{tab:Ablation-LoD}
  \centering
  \tabcolsep=0.2cm
  \resizebox{0.9\textwidth}{!}{%
  \begin{tabular}{@{}ll|llll@{}}
    \toprule
    In-Frustum & Distance Interval (m) & SSIM$\uparrow$ & PSNR$\uparrow$ & LPIPS$\downarrow$ & FPS$\uparrow$ \\
    \midrule
    block-wise & [0,200],[200,400],[400,$\infty$] & 0.855 & 27.32 & 0.229 & 53.7 \\
    point-wise & [0,200],[200,400],[400,$\infty$] & 0.849 & 27.18 & 0.239 & 30.3 \\
    block-wise & [0,150],[150,300],[300,$\infty$] & 0.848 & 27.16 & 0.242 & \textbf{57.6} \\
    block-wise & [0,250],[250,500],[500,$\infty$] & \textbf{0.858} & \textbf{27.39} & \textbf{0.223} & 46.4 \\
    \bottomrule
  \end{tabular}
  }
\end{table}

The quantitative ablation on LoD strategy is shown in \cref{tab:Ablation-LoD}. Comparing the first and the second row, the computation burden introduced by the point-wise strategy leads to the considerably worse real-time performance. Its rendering quality is on par with the smaller interval setting, i.e. third row. Comparing the first row with the third and the fourth row, it can be observed that with larger intervals for LoD 2 and LoD 1, the rendered results would contain richer details, but downgraded speed. And first row arrives at a balance, which is adopted as the standard setting.

%% file: sections/5_conclusions.tex
\section{Conclusions}
In this work, we present CityGS, which successfully realizes real-time large-scale scene reconstruction with high fidelity. Through the blocking and LoD strategy tailored for Gaussian geometry, we obtain state-of-the-art rendering fidelity on mainstream benchmarks, while significantly reducing time costs when rendering drastically different scales of the same scene. However, the hidden static scene assumption limits its generalization ability. Training with the combination of drastically different views such as aerial and street views also degrades instead of boosting the performance of CityGS. The inner mechanism deserves to be further explored and well-resolved. 

% there are still some issues that we aim to address in the future: 1) CityGS assumes the scene to be static. But most images are captured under different weather conditions and contain transient objects. Thus the generalization ability of CityGS can be improved. 2) Training with the combination of street view and aerial view remains challenging. The street view complements the aerial view with near-ground details, but performance indeed degrades when trained together. The inner mechanism deserves to be explored and well-resolved. 

%% file: sections/6_appendix_arXiv.tex
\setcounter{equation}{0}
\setcounter{table}{0}
\setcounter{figure}{0}
\setcounter{section}{0}
\renewcommand\theequation{S\arabic{equation}}
\renewcommand\thefigure{S\arabic{figure}}
\renewcommand\thetable{S\arabic{table}}
\renewcommand\thesection{\Alph{section}}

% \section*{Supplementary Material}
\section{Additional Experimental Results}
\label{ssec:More Exp}

\subsection{Additional Quantitative Comparison}
\label{ssubsec:More quantitative}
The quantitative comparisons across datasets \textit{MatrixCity}, \textit{Residence}, \textit{Rubble}, \textit{Building}, and \textit{Sci-Art} are presented in \cref{stab:SOTA}. We not only provide the performance of the CityGS with no LoD as done in Sec. 4.2 of our main paper, but also presents the standard CityGS for reference. It can be observed that our approach outperforms others in terms of \textbf{SSIM} and \textbf{LPIPS} among all the datasets, and achieves the highest \textbf{PSNR} on \textit{MatrixCity}, \textit{Rubble}, and \textit{Building}. The relatively weaker \textbf{PSNR} of \textit{Sci-Art} and \textit{Residence} is mainly attributed to the appearance variations across views in these dataset. We leave solving this issue for future works. Thanks to the superior efficiency of 3DGS, we have achieved much faster speed than previous state-of-the-art even without LoD. 

\begin{table}
  \caption{Quantitative Comparison on five large-scale scene datasets. The '-' symbol indicates Mega-NeRF \cite{turki2022mega} and Switch-NeRF \cite{zhenxing2022switch} were not evaluated on \textit{MatrixCity} due to difficulties in adjusting its training configurations beyond the provided, resulting in poor performance on this dataset. The best results of each metric are in \textbf{bold}.
  }
  \label{stab:SOTA}
  \centering
  \tabcolsep=0.08cm
  \resizebox{0.99\textwidth}{!}{%
  \begin{tabular}{@{}l|llll|llll|llll|llll|llll@{}}
    \toprule
     & \multicolumn{4}{c|}{MatrixCity} & \multicolumn{4}{c|}{Residence} & \multicolumn{4}{c|}{Rubble} & \multicolumn{4}{c|}{Building} & \multicolumn{4}{c}{Sci-Art}\\
    \midrule
    Metrics & SSIM$\uparrow$ & PSNR$\uparrow$ & LPIPS$\downarrow$ & FPS$\uparrow$ \ & SSIM$\uparrow$ & PSNR$\uparrow$ & LPIPS$\downarrow$ & FPS$\uparrow$ \ & SSIM$\uparrow$ & PSNR$\uparrow$ & LPIPS$\downarrow$ & FPS$\uparrow$ \ & SSIM$\uparrow$ & PSNR$\uparrow$ & LPIPS$\downarrow$ & FPS$\uparrow$ \ & SSIM$\uparrow$ & PSNR$\uparrow$ & LPIPS$\downarrow$ & FPS$\uparrow$\\
    \midrule
    MegaNeRF \cite{turki2022mega} & - & - & - & - & 0.628 & 22.08 & 0.489 & <0.1 & 0.553 & 24.06 & 0.516 & <0.1 & 0.547 & 20.93 & 0.504 & <0.1 & 0.770 & 25.60 & 0.390 & <0.1 \\
    Switch-NeRF \cite{zhenxing2022switch} & - & - & - & - & 0.654 & \textbf{22.57} & 0.457 & <0.1 & 0.562 & 24.31 & 0.496 & <0.1 & 0.579 & 21.54 & 0.474 & <0.1 & 0.795 & \textbf{26.61} & 0.360 & <0.1 \\
    GP-NeRF \cite{xu2023grid} & 0.611 & 23.56 & 0.630 & 0.15 & 0.661 & 22.31 & 0.448 & 0.31 & 0.565 & 24.06 & 0.496 & 0.40 & 0.566 & 21.03 & 0.486 & 0.42 & 0.783 & 25.37 & 0.373 & 0.34 \\
    3DGS$^\dagger$ \cite{kerbl20233d} & 0.735 & 23.67 & 0.384 & 35.9 & 0.791 & 21.44 & 0.236 & \textbf{62.1} & 0.777 & 25.47 & 0.277 & 47.8 & 0.720 & 20.46 & 0.305 & \textbf{45.0} & 0.830 & 21.05 & 0.242 & \textbf{72.2}\\
    CityGS(no LoD) & \textbf{0.865} & \textbf{27.46} & \textbf{0.204} & 21.6 &  \textbf{0.813} & 22.00 & \textbf{0.211} & 32.7 & \textbf{0.813}& \textbf{25.77} & \textbf{0.228} & 43.9 & \textbf{0.778} & 21.55 & \textbf{0.246} & 24.3 &  \textbf{0.837} & 21.39 & \textbf{0.230} & 56.1 \\
    CityGS & 0.855 & 27.32 & 0.229 & \textbf{53.7} & 0.805 & 21.90 & 0.217 & 41.6 & 0.785 & 24.90 & 0.256 & \textbf{52.6} & 0.764 & \textbf{21.67} & 0.262 & 37.4 & 0.833 & 21.34 & 0.232 & 64.6 \\
    \bottomrule
  
  \end{tabular}
  }
\end{table}

\begin{table}
    \caption{Ablation on block numbers and SSIM threshold $\varepsilon$. The experiment is conducted on the \textit{Rubble} dataset. The first row is the performance of coarse global Gaussians prior mentioned in Sec. 3.2 of main paper, and thus has no block number or $\varepsilon$ setting. MEAN denotes an average number of 
    assigned training poses among divided blocks. The best performances are in \textbf{bold}.}
    \label{stab:Ablation-Training}
    \centering
    \tabcolsep=0.4cm
    \resizebox{0.7\textwidth}{!}{%
    \begin{tabular}{@{}cc|llll@{}}
    \toprule
    $\varepsilon$ & \#Blocks & SSIM$\uparrow$ & PSNR$\uparrow$ & LPIPS$\downarrow$ & MEAN \\
    \midrule
     % n/a & n/a & 0.770 & 24.83 & 0.284 & n/a\
     0.1 & $2\times2$ & 0.762 & 24.48 & 0.285 & 835\\
     0.1 & $3\times3$ & 0.806 & 25.41 & 0.238 & 552 \\
     0.1 & $4\times4$ & 0.804 & 25.45 & 0.241 & 432 \\
     % 0.1 & $5\times5$ & \textbf{0.814} & 25.70 & \textbf{0.228} & 355 & 115 \\
     0.12 & $3\times3$ & \textbf{0.813} & \textbf{25.77} & 0.231 & 469 \\
     0.14 & $3\times3$ & 0.812 & 25.43 & \textbf{0.229} & 415 \\
    \bottomrule
  \end{tabular}
  }
\end{table}

Besides, as shown in \cref{stab:SOTA}, LoD significantly improves efficiency, especially for extremely large-scale scenes such as \textit{MatrixCity}. Compared with our CityGS, the 3DGS$^\dagger$ \cite{kerbl20233d} possesses faster speed but significantly lower rendering quality. The main reason is that the original 3DGS requires sufficiently large iterations and memory to optimize the whole scene with thousands of images. Bounded by computation resources, the capacity of the trained original 3DGS is too limited to represent the whole large-scale scene well.

\subsection{Additional Ablations}
\label{ssubsec:More ablation}
We also explored the influence of hyper-parameters in training, namely block number and data assignment threshold $\varepsilon$ mentioned in Sec. 3.2 of main paper. Here, we control the overall finetuning iterations to be $9\times32000$. As shown in \cref{stab:Ablation-Training}, as the block number grows, the average data assigned to blocks decreases. We achieve optimal results around 3 × 3 partitions. The $\varepsilon$ also controls data assignment. As $\varepsilon$ grows, the average poses assigned decreases. And if it is too high, many necessary training data will be lost, thus leading to lower PSNR performance.

\begin{table}
  \caption{Detailed Parameter Setting. In \textbf{training}, the foreground area for contraction is bounded by $p_{\min}=(x_{\min},y_{\min},z_{\min})$ and $p_{\max}=(x_{\max},y_{\max},z_{\max})$. Here we take the height dimension as $z$. From the bird's eye view, the longest side is set as the x-axis, while the shortest is the y-axis. $z$ bound is set as the minimum and maximum position of all Gaussians, and thus not included. The block dimension along the x-axis and y-axis is denoted as \#Blocks, and $\varepsilon$ is SSIM threshold. In \textbf{rendering}, the Distance Interval decides detail level assignment.}
  \label{stab:param}
  \centering
  \tabcolsep=0.2cm
  \resizebox{0.99\textwidth}{!}{%
  \begin{tabular}{@{}l|llllll|l@{}}
    \toprule
    Dataset & $x_{\min}(m)$ & $y_{\min}(m)$ & $x_{\max}(m)$ & $y_{\max}(m)$ & \#Blocks & $\varepsilon$ & Distance Interval (m) \\
    \midrule
    MatrixCity & -350 & -400 & 450 & 200 & $6 \times 6$ & 0.05 & [0,200],[200,400],[400,$\infty$]\\
    Rubble & -50 & -5 & 50 & -135& $3 \times 3$ & 0.12 & [0,100],[100,200],[200,$\infty$]\\
    Building & -140 & 250 & -10 & 0 & $5 \times 4$ & 0.1 & [0,100],[100,200],[200,$\infty$]\\
    Residence & -270 & -25 & 60 & 175 & $5 \times 4$ & 0.08 & [0,250],[250,500],[500,$\infty$]\\
    Sci-Art & -205 & -110 & 90 & 55 & $3 \times 3$ & 0.05 & [0,250],[250,500],[500,$\infty$]\\
    \bottomrule
  \end{tabular}
  }
\end{table}

\section{Detailed Parameter Setting}
\label{ssec:Params}

We present the specific hyper-parameter configurations for each dataset in \cref{stab:param}. The roles of training parameters are detailed in Sec. 3.2 of the main paper, while the role of rendering parameter Distance Interval is specified in Sec. 3.3 of the main paper. Note that for LoD on datasets except for \textit{MatrixCity}, we use three detail levels of compression rate 60\%, 50\%, and 40\%.

\begin{figure}
\centering
\includegraphics[width=0.99\textwidth]{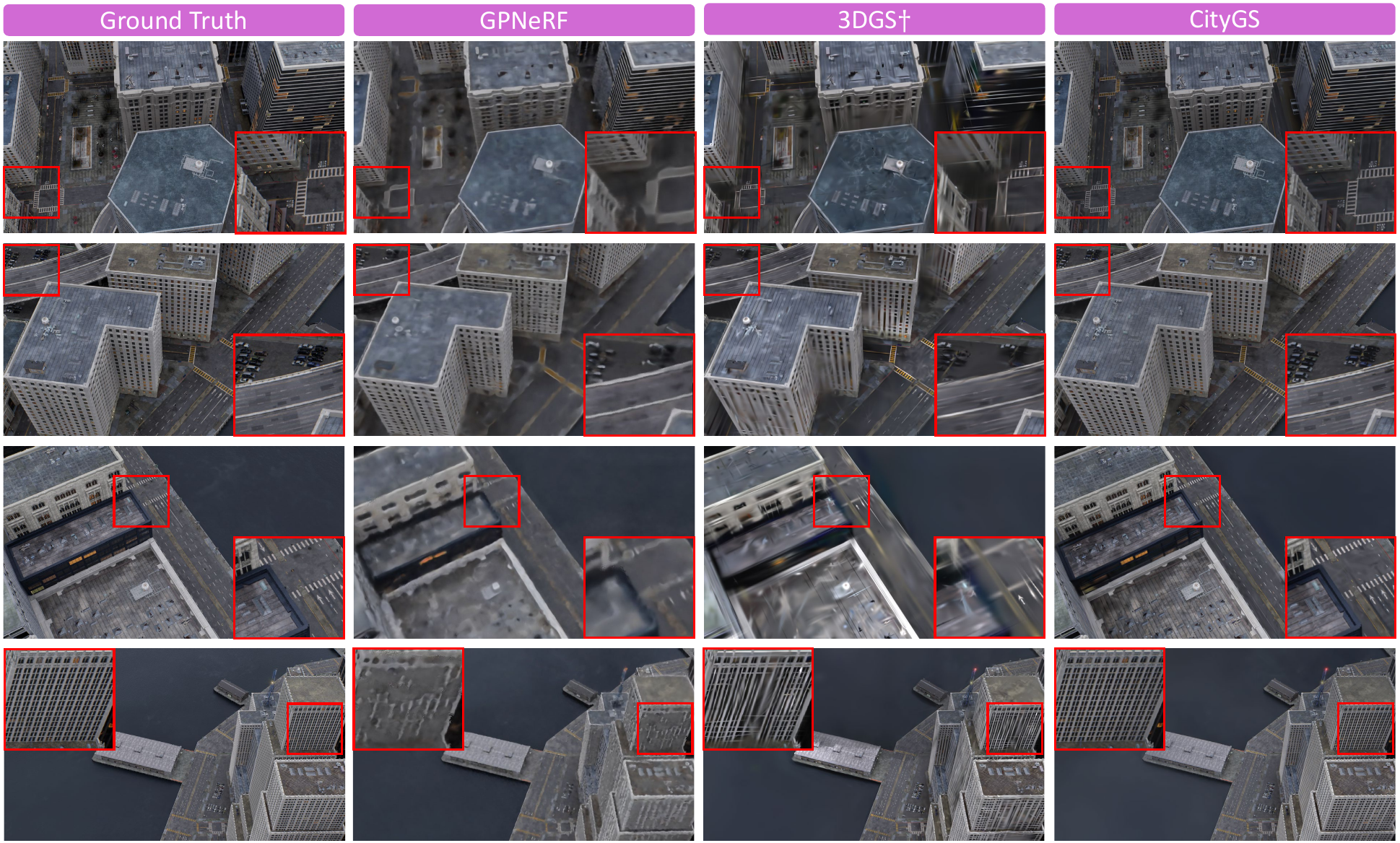}
\caption{More qualitative comparison with SOTA methods on \textit{MatrixCity} dataset.}
\label{sfig:Vis-Sim-Supp}
\end{figure}

\section{More Visualization on MatrixCity Dataset}
\label{ssec:More Vis}
In this section, we provide additional qualitative comparisons using the \textit{MatrixCity} dataset, depicted in \cref{sfig:Vis-Sim-Supp}. Our results showcase the superior reconstruction quality of intricate details, including crowded cars and crosswalks.  The remarkable enhancement in visual fidelity compared with other methods sufficiently illustrates the superiority of our CityGS. 

\section{Qualitative Validation on Concatenated Fusion}
\label{ssec:continuity}
\begin{figure}
\centering
\includegraphics[width=0.99\textwidth]{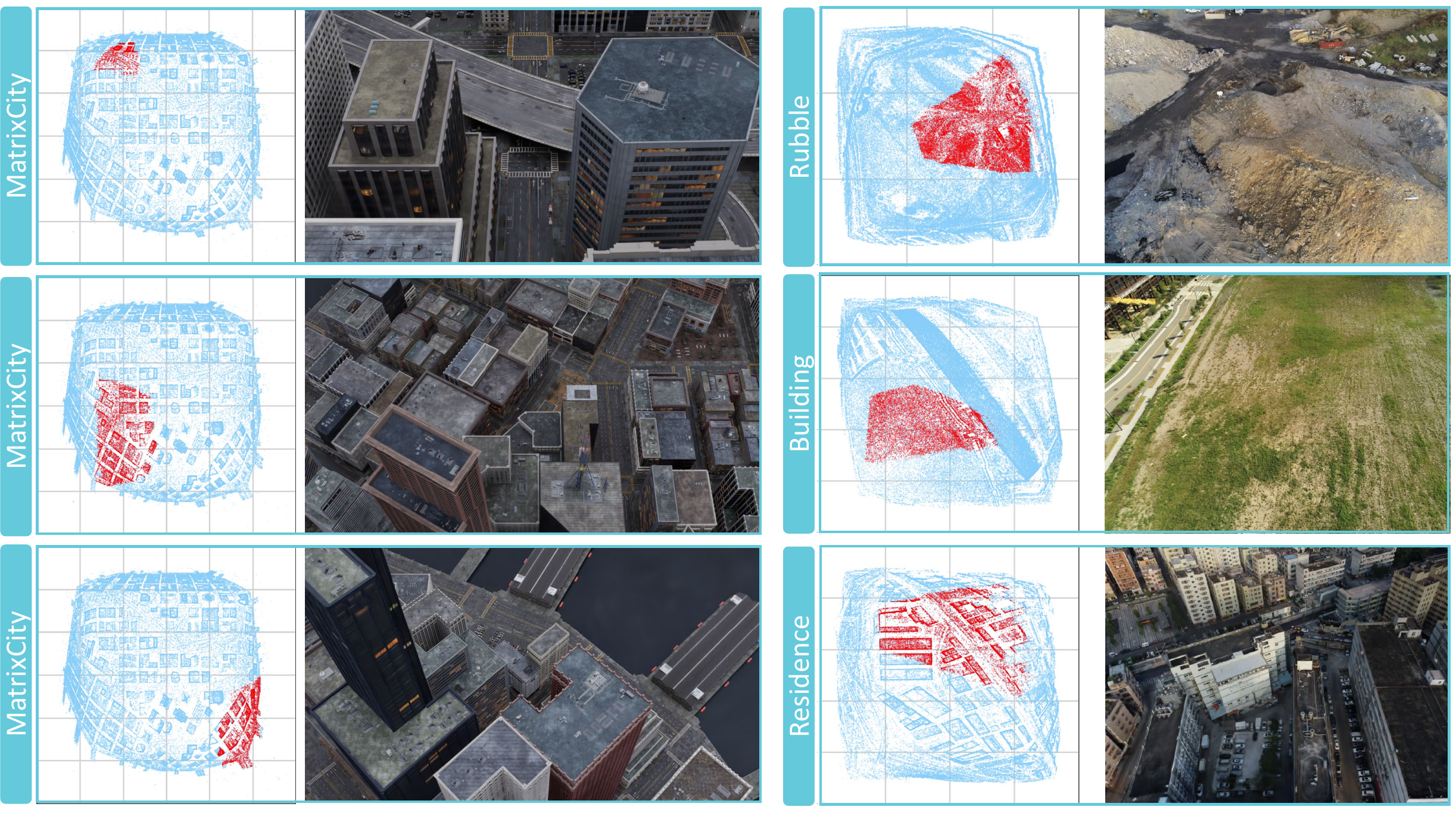}
\caption{Qualitative validation of boundary continuity on both synthetic and real datasets when visible Gaussians across multiple blocks. Each subfigure illustrates point distribution under contracted space on the left and rendered image on the right. For the point distribution, \textcolor[RGB]{133,208,247}{blue} points denote overall Gaussians, while the \textcolor[RGB]{255,0,0}{red} points denote visible Gaussians. The \textcolor[RGB]{211,211,211}{grey} grid depicts block partition under contracted space. }
\label{sfig:Continuity}
\end{figure}

To validate the effectiveness of the concatenated fusion strategy, we perform rendering at the viewpoints where the visible area spans multiple blocks. The Gaussians utilized here come from direct concatenation of fine-tuned Gaussians of corresponding blocks. As depicted in \cref{sfig:Continuity}, rendering from a specific viewpoint may involve four or more blocks. Despite that, the rendered images exhibit no discernible discontinuities, showcasing smooth boundary transitions facilitated by our coarse global Gaussian prior, as discussed in Sec. 3.2 of our main paper.

\section{Scene Manipulation}
\label{ssec:-Manipulation}

\begin{figure}
\centering
\includegraphics[width=0.99\textwidth]{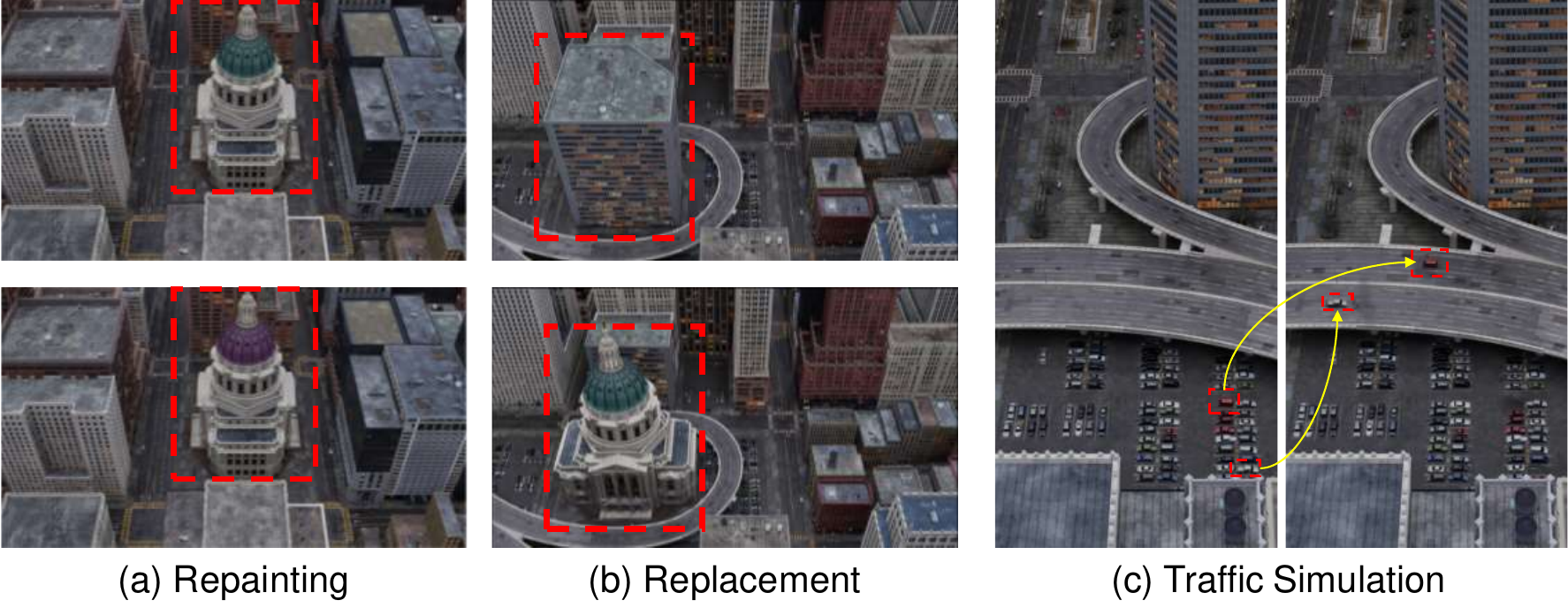}
\caption{Illustration of city scene manipulation driven by explicit representation of CityGS. In Part (a), the dome of the original building in the first row is repainted to the desired color shown in the second row. In Part (b), the building of the first row is removed and replaced with the one shown in the second row. In Part (c), the cars parked at locations shown in the left image are moved to the positions shown in the right image, so as to simulate the required traffic conditions. NeRF-based methods struggle to realize such manipulation.}
\label{sfig:scene manipulation}
\end{figure}

For the implicit representation of NeRF-based methods, it is hard to explain the correspondence between network parameters and scene structure. However, since we can reconstruct the explicit city representation with relatively high geometric precision in CityGS, the geometric and appearance distribution can be manipulated as desired. The demos are shown in \cref{sfig:scene manipulation}. The appearance of a specified part of a building can be transformed to the desired style. It is also possible to delete a building and replace it with another one. By placing cars or pedestrians, the pre-defined traffic conditions can be simulated in the city. These demos indicate potential real-time and interactive application of CityGS.